\documentclass[sigconf,balance=true]{acmart}
\AtBeginDocument{%
  \providecommand\BibTeX{{%
    \normalfont B\kern-0.5em{\scshape i\kern-0.25em b}\kern-0.8em\TeX}}}

\setcopyright{rightsretained}
\acmYear{2021}
\acmISBN{}
\acmDOI{}
\acmYear{2021}
\acmISBN{}
\acmDOI{}
\acmConference[PatentSemTech]{2nd Workshop on Patent Text Mining and Semantic Technologies (PatentSemTech2021) co-located with the 44th International ACM SIGIR Conference on Research and Development in Information Retrieval}{July 15th, 2021}{online}
\makeatletter
\renewcommand{\@copyrightpermission}{}
\renewcommand{\@copyrightowner}{for this paper by its authors. Use permitted under Creative Commons License Attribution 4.0 International (CC BY 4.0). CEUR Workshop Proceedings (CEUR-WS.org)}
\makeatother
\usepackage{algorithmic}
\usepackage{graphicx}
\usepackage{textcomp}
\usepackage{xcolor}

\usepackage{listings}
\usepackage[utf8]{inputenc}
\usepackage[shortlabels]{enumitem}

\begin{document}

\title{Prior Art Search and Reranking \\for Generated Patent Text}


\author{Jieh-Sheng Lee}
\orcid{0000-0002-0990-6170}
\authornote{Admitted in New York and passed the USPTO patent bar exam.}
\email{d04922013@csie.ntu.edu.tw}
\affiliation{
  \institution{National Taiwan University}
  \streetaddress{}
  \city{Taipei}
  \country{Taiwan}
}

\author{Jieh Hsiang}
\email{hsiang@csie.ntu.edu.tw}
\affiliation{
  \institution{National Taiwan University}
  \streetaddress{}
  \city{Taipei}
  \country{Taiwan}
}


\begin{abstract}
Generative models, such as GPT-2, have demonstrated impressive results recently. A fundamental question we would like to address is: where did the generated text come from? This work is our initial effort toward answering the question by using prior art search. The purpose of the prior art search is to find the most similar prior text in the training data of GPT-2. We take a reranking approach and apply it to the patent domain. Specifically, we pre-train GPT-2 models from scratch by using the patent data from the USPTO. The input for the prior art search is the patent text generated by the GPT-2 model. We also pre-trained BERT models from scratch for converting patent text to embeddings. The steps of reranking are: (1) search the most similar text in the training data of GPT-2 by taking a bag-of-words ranking approach (BM25), (2) convert the search results in text format to BERT embeddings, and (3) provide the final result by ranking the BERT embeddings based on their similarities with the patent text generated by GPT-2. The experiments in this work show that such reranking is better than ranking with embeddings alone. However, our mixed results also indicate that calculating the semantic similarities among long text spans is still challenging. To our knowledge, this work is the first to implement a reranking system to identify retrospectively the most similar inputs to a GPT model based on its output.
\end{abstract}






\maketitle

\section{Introduction}

Generative models based on Deep Learning techniques have shown significant progress in recent years. A long-term objective of our research is to evaluate the novelty of the text produced by the generative models in the patent domain. Before evaluating the novelty, a prerequisite is to identify the closest prior arts. The scope of the prior arts is the patent text used for training the generative models. From the perspective of system implementation, the approach this paper took is to integrate ``patent text generation'' and ``prior art search.'' The purpose of this paper is to fulfill the prerequisite for identifying prior arts so that the novelty of the generated patent text can be evaluated in the future. Assuming that the opposite of the novelty in text generation is memorizing training text, the pre-training of the GPT-2 generative models in~\cite{gpt2_Radrof01} is indicative of such novelty. In the paper, the authors observed some text memorizing behavior in their models on longer strings that are repeated many times in the dataset. The authors quantified how often exact memorization shows up in the generated text by measuring the percentage in the 8-gram overlap. According to the authors, most samples have less than 1\% overlap, including over 30\% of samples with no overlap. Such results indicate that GPT-2 models can generate novel text relatively well.

In the patent domain, the authors in~\cite{jiehsheng03} applied such GPT-2 model to generate patent claims. The authors proposed an idea called ``Augmented Inventing,'' aiming to help inventors conceive new patents in a better way. Since patent claims are generally longer than ordinary sentences, the authors proposed a ``span-based'' approach to decompose a long patent text into multiple shorter text spans. The authors also proposed an idea called ``auto-complete'' function to generate patent text on a span basis. From a legal perspective, such a function will be valuable if it can generate something new and meet (at least) the ``novelty'' requirement in patent laws. However, for a generative model to meet the legal requirement, a fundamental question is to calculate the similarity between generated patent text and prior patents. In~\cite{jiehsheng03}, the GPT-2 model can generate plausible patent claims in surface form, but it is unclear how novel the patent text is. To address the problem, the authors proposed a dual-Transformer framework (using one Transformer to measure the other Transformer), and they tried to measure the quality of patent text generation (by span relevancy in~\cite{jiehsheng02} and by semantic search in~\cite{jiehsheng08}). Despite these efforts, measuring the novelty in patent text generation remains an open problem. 

From a different perspective, building a generative patent model to augment inventors might be the beginning of the era of human-machine co-inventing or meta-inventing (inventing how to invent). In such an era, measuring the novelty created by the generative model will be an essential function. To measure the novelty, it is required to compare the output of the model and its inputs. In this work, our implementation scope is to compare the generated patent text with the original patent text in the training dataset. Since the training dataset is large, in order to narrow the scope of comparison, it is required to identify the most similar prior text in the training dataset. Therefore, our implementation is to build such a prior art search system. We found that reranking is a practical way to make the search more effective. As proof of concept, we limit the data scope in this work to granted patents only. The prior art search is also limited to finding the most relevant text in span-based fashion. How to aggregate the similarities of multiple text spans into a longer sentence or a paragraph is another topic in the future.






\section{Related Work}

Our prior art search's main challenge is how to calculate the semantic similarity between two patent text spans. In the past, most of the prior art searches were performed at the word level, such as keywords or phrases. For example, the authors in~\cite{Dhondt_Verberne_Weber_Koster_Boves_2012} found that combining unigrams and PoS-filtered skipgrams leads to a significant improvement in classification scores over the unigram baseline. 
In recent years, researchers moved toward neural network models and embeddings for the semantic search of longer text. For example, in~\cite{Risch2019DomainspecificWE}, the authors utilized domain-specific word embeddings for patent classification. In~\cite{logeswaran2018an}, the authors proposed ``Quick Thought'' to represent a sentence in a fixed-length vector. The scheme is similar to the skip-gram method in Word2Vec~\cite{word2vec} by escalating the idea from word level to sentence level. Another line of development is based on new neural architectures, such as Transformer~\cite{Vaswani01}. Notably, BERT~\cite{devlin-etal-2019-bert} and RoBERTa~\cite{RoBERTa} set a new state-of-the-art performance on sentence-pair regression tasks, e.g., semantic textual similarity (STS). According to~\cite{SentenceBERT}, however, BERT is unsuitable for semantic similarity search on a large scale. For example, finding the most similar pairs in a collection of 10,000 sentences requires about 50 million inference computations (65 hours) with BERT. The authors in~\cite{SentenceBERT} proposed a modification of the pre-trained BERT model to use siamese and triplet network structures. Their model called ``Sentence-BERT'' can derive semantically meaningful sentence embeddings to be compared by using cosine similarity. Significantly it reduces the effort for finding the most similar pair from 65 hours with BERT/RoBERTa to about 5 seconds while maintaining the accuracy from BERT, according to~\cite{SentenceBERT}. Specific to the patent domain in~\cite{jiehsheng05}, the authors showed that embedding could be a better metric than conventional ROGUE (word-based) for measuring semantic similarity. The metric for measuring embeddings in~\cite{jiehsheng05} is based on the Universal Sentence Encoder (USE)~\cite{universal-sentence-encoder} without any fine-tuning. 

A further line of development is to combine both word level and embedding level. For example, NBoost~\cite{nboost} can deploy Transformer models to improve the relevance of search results on conventional word-based search engines, such as Elasticsearch using BM25. According to~\cite{nboost}, NBoost works like a proxy between users and Elasticsearch. It leverages fine-tuned models to produce domain-specific results. In a search request, the user sends a query to NBoost. Then, NBoost asks for results from Elasticsearch, picks the best ones based on the fine-tuned model, and returns its final results to the user. Specifically, if a user asks for 10 results, NBoost can increase the number of requests for Elasticsearch to produce 100 records (word-based) and then pick the best 10 results (embedding-based). Such a technique is called reranking.

\section{Approach}

\subsection{Semantic Search with Reranking}
\label{subsection:semantic_search}

Compared with contextualized word embeddings, we found the research in sentence embeddings more challenging and less explored. For example, the USE model in~\cite{universal-sentence-encoder} is publicly available, but the code for pre-training or fine-tuning is not. Without fine-tuning with domain-specific data, a model could deviate from a downstream task and fail to perform well in the specific domain. Our experience found that the USE model alone without fine-tuning is not satisfactory for having a useful metric to measure the semantic similarity in patent spans. We also found that with a BERT model, even pre-trained with patent corpus, the false-positive rate of the semantic similarity based on BERT embeddings is still high. The Sentence-BERT in~\cite{SentenceBERT} might be a solution to these problems. However, if we would like to take the Sentence-BERT approach, an obstacle will be data. Sentence-BERT requires both positive and negative examples to learn the similarity function. In this work, all of the data from USPTO are positive examples. As for how to prepare negative examples in the future, the PatentMatch dataset for training a binary text pair classifier in~\cite{risch2020patentmatch} can be a reference.

Since none of those above (USE, BERT, and Sentence-BERT) is a viable option, we resorted to the reranking idea demonstrated by NBoost. Besides, we found that the first author of~\cite{SentenceBERT} proposes a similar view on his GitHub repository. According to the author, a bag-of-words search, such as BM25, can have a higher recall, but its precision is lower. Conversely, the embedding-based search can have higher precision, but its recall is lower. Therefore, for having both the higher recall and the higher precision, a reranking strategy is to perform the word-based ranking upfront for a higher recall and then perform the embedding-based ranking for higher precision. It is noted that, in our initial experiments, we found that ranking based on embeddings first and reranking based on words later does not perform well. In such a configuration, the false-positive rate in the ranking of embeddings is too high.

\subsection{System Architecture}

\begin{figure}[t]
    \centering
    \includegraphics[width=\linewidth]{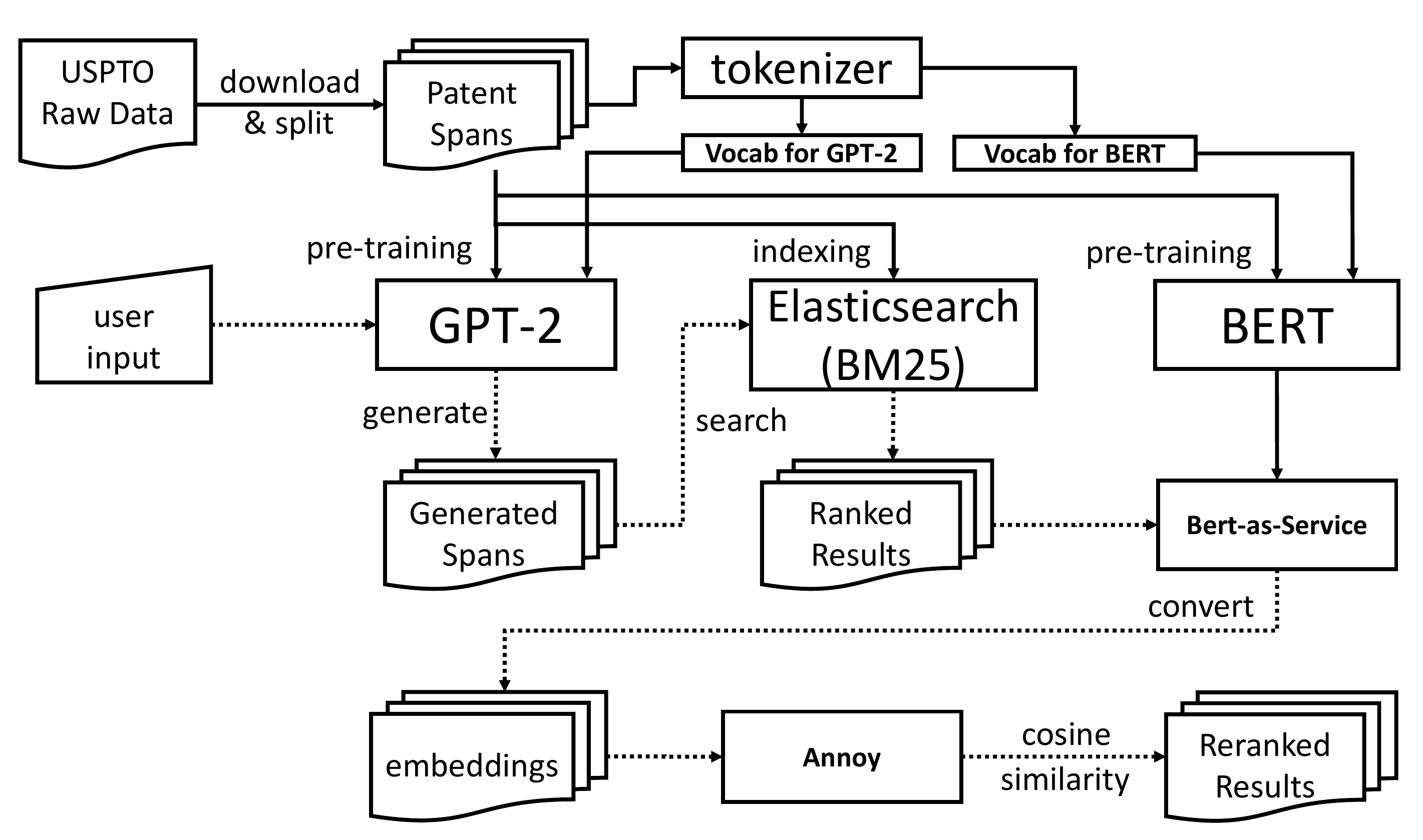}
    \caption{System Architecture}
    \label{fig:architecture}
\end{figure}

This section explains the overall architecture of our implementation and the function/data flows in the architecture. For details, section~\ref{section:data} will cover the data part and its preprocessing, and section~\ref{subsection:github} will provide the code repositories we leveraged from others. Fig.~\ref{fig:architecture} shows our system architecture. The upper portion of the figure (stage 1 of our implementation) represents what we have to build before a user can trigger GPT-2 for text generation. The function flows in this upper portion are depicted in solid lines. The bottom portion of the figure (stage 2 of our implementation) shows the function flows (in dotted lines) for ranking (BM25) and reranking (BERT embeddings). 

At stage 1, we download raw data from the USPTO and split them into patent spans. The patent spans are fed into Elasticsearch (with default settings) for indexing so that we can query them based on BM25 ranking. Before pre-training GPT-2 and BERT from scratch, we built their vocabulary files. The details of pre-trainings GPT-2 and BERT are provided in sections~\ref{subsection:gpt2_implementation} and~\ref{subsection:bert_implementation}, respectively. At stage 2, a user provides the input text to GPT-2 and the parameters of GPT-2 inferencing. GPT-2 generates a patent text span based on these settings. For reranking, the generated patent span first goes to Elasticsearch to obtain the most relevant prior patent spans based on BM25. The ranked prior patent spans then go to Bert-as-Service~\cite{xiao2018bertservice} and convert to BERT embeddings. Next, the embeddings go to Annoy~\cite{annoy0001} for reranking based on cosine similarity. Lastly, the final and reranked embeddings are decoded back to patent spans in text format and shown to the user. In our experiments, based on the same user's input, we let GPT-2 generate multiple patent text spans to collect both positive and negative reranking results based on each of the multiple patent text spans.

\section{Data}
\label{section:data}

\subsection{Data source}

In~\cite{jiehsheng01}, the authors use the patent datasets on BigQuery provided by Google~\cite{bigquery}. Although it is flexible to manipulate the data by SQL statements, we found that the data provided by BigQuery are not updated frequently. 
We turned to the USPTO PatentsView~\cite{patentsview} for bulk download files and more updates. At the moment of this writing, the latest version of the ``patent.tsv.zip'' file is dated as 2020-03-31.
For incremental download instead of bulk download, the USPTO Open Data Portal~\cite{opendataportal} can be another choice. 
The raw data provided by the PatentsView and the Open Data Portal are plain text in TSV or XML format. The downside of using such raw text is the extra efforts on data preprocessing, compared with the flexibility of SQL statements in BigQeury. Practitioners need to consider the tradeoff between flexibility and data frequency. We opt for more frequently updated data in this work.  

\subsection{Datasets for GPT-2}
\label{subsection:dataset_for_gpt2}

During data preprocessing, we follow the span-based approach in~\cite{jiehsheng03} and follow the ``structural metadata'' and ``metadata mapping'' approaches in~\cite{jiehsheng05}. The structural metadata in~\cite{jiehsheng05} is defined to include patent title, abstract, independent claim, and dependent claim. According to the authors, it is a mechanism to control what kind of patent text to generate. Regarding metadata mapping, it is a mechanism to guide GPT-2 for generating from one kind of patent text to another. What we differ from~\cite{jiehsheng05} are: (1) we add new tags for patent drawing descriptions, (2) we add \lstinline{<|dep|>} for dependent claims, (3) we remove the proposed ``backward'' tags because backward text generation is not required in this work. Table \ref{tab:specialtag} shows our special tags for structural metadata and mappings between metadata. We found the span-based approach helpful for splitting long claims into short text spans. However, for patent abstracts, such a span-based approach may not apply. If an abstract's text has been taken verbatim from a claim, the span splitting mechanism may apply. If not, there might be no span to split in a sentence. When no span is found in the abstract, we split a patent abstract into multiple sentences instead. Collectively the split sentences or spans are treated the same way in our data processing, and we refer to both of them as ``span'' in this work. 

\begin{table}[t]
  \caption{Special Tags \& Mappings}
  \label{tab:specialtag}
  \begin{center}
  \begin{tabular}{|c|c|c|}
    \hline
    \multicolumn{3}{|c|}{\textbf{Tags for Metadata}} \\
    \hline
    \textbf{metadata} & \textbf{prefix} & \textbf{appendix} \\
    \hline
    title & \textless{\textbar}start\_of\_title\textbar\textgreater & \textless{\textbar}end\_of\_title\textbar\textgreater \\
    \hline
    abstract & \textless{\textbar}start\_of\_abstract\textbar\textgreater & \textless{\textbar}end\_of\_abstract\textbar\textgreater \\
    \hline
    figure & \textless{\textbar}start\_of\_figure\textbar\textgreater & \textless{\textbar}end\_of\_figure\textbar\textgreater \\
    \hline
    independent claim & \textless{\textbar}start\_of\_claim\textbar\textgreater & \textless{\textbar}end\_of\_claim\textbar\textgreater \\
    \hline
    dependent claim & \textless{\textbar}dep\textbar\textgreater \textless{\textbar}start\_of\_claim\textbar\textgreater & \textless{\textbar}end\_of\_claim\textbar\textgreater \\
    \hline    
    span / sentence & (n/a) & \textless{\textbar}span\textbar\textgreater \\
    \hline    
    \hline
    \multicolumn{3}{|c|}{\textbf{Metadata Mappings}} \\
    \hline
    \textbf{metadata 1} & \textbf{mapping} & \textbf{metadata 2} \\
    \hline
    title & \textless{\textbar}title2abstract\textbar\textgreater & abstract \\
    \hline
    abstract & \textless{\textbar}abstract2title\textbar\textgreater & title \\
    \hline
    claim & \textless{\textbar}claim2abstract\textbar\textgreater & abstract \\
    \hline
    abstract & \textless{\textbar}abstract2claim\textbar\textgreater & claim \\
    \hline
    title & \textless{\textbar}title2figure\textbar\textgreater & figure \\
    \hline
    figure & \textless{\textbar}figure2title\textbar\textgreater & title \\
    \hline
  \end{tabular}
  \end{center}
\end{table}

Based on the approaches as mentioned above, the actual pipeline to build the datasets for GPT-2 includes: (1) downloading the raw TSV file from USPTO PatentsView and splitting them into smaller files, (2) extracting text based on metadata (e.g., title, abstract, etc.) and uploading them to the Elasticsearch server, (3) retrieving patent text from the Elasticsearch server, adding special tags to them, and saving them in text format (4) converting the text files in the previous step to TFRecord format for Tensorflow code. In step (1), the text data from USPTO PatentsView is about 48.7G (version: 2019-10-08). Such a corpus is larger than the WebText corpus (40G) used by OpenAI for GPT-2 pre-training. In step (4), the total number of tokens is 32.3B (32,398,927,872). 


%

By concatenating all of the text with special tags, the total amount of data reaches 180G. Due to resource constraints, we did not concatenate a dependent claim with its corresponding independent claim. Independent claims are generally much longer than their dependent claims. If our training data capture such claim dependency for all dependent claims, e.g., ``(claim 1) $<$\textbar dep\textbar$>$ (claim 2)'' and ``...$<$\textbar abstract2claim\textbar$>$$<$\textbar start\_of\_claim\textbar$>$ (claim1+claim2)'' (some special tags omitted for clarity), it is possible that the total amount of text data may exceed 570G (the size of text data for training GPT-3~\cite{gpt-3}). We leave such an experiment for future researchers. It is also noted that the $<$\textbar figure2title\textbar$>$ mapping in Table \ref{tab:specialtag} does not exist in our training data. We reserve this mapping for testing whether it is possible for GPT-2 models to do the same few-shot learning in GPT-3. 

\subsection{Datasets for BERT}

According to BERT's code repository~\cite{bert_github}, the input for pre-training is a plain text file having one sentence per line. Consecutive lines are the actual sentences for the "next sentence prediction" task. Documents are delimited by empty lines. The final datasets contain serialized text in TFRecord file format. In our case, we follow the format and prepare the plain text file with one span or sentence per line. We did not add our special tags or metadata mappings to the text file because such annotations are designed for GPT-2 only. The total number of words serialized in our training data is 6.8 billion words. It is larger than the 3.3 billion word corpus (BooksCorpus with 800M words and English Wikipedia with 2,500M words) for pre-training the official BERT model.

\subsection{Data for Elasticsearch server}

The purpose of the Elasticsearch server in our data pipeline is two-folded. First, it provides the ranking mechanism based on a bag-of-words approach. For example, we can query the top \textit{n} records (e.g., 100) based on BM25. Second, the Elasticsearch server is convenient for us to aggregate various patent text from different raw files. Such aggregation replaces the BigQuery and SQL statements in~\cite{jiehsheng01}. In step (2) of the data pipeline in~\ref{subsection:dataset_for_gpt2}, we split the patent text into spans or sentences and upload them to the Elasticsearch server. The total number of records in Elasticsearch is 343,987,632, and they occupy 59.7GB.

\section{Implementation \& Experiments}

\subsection{GitHub repositories}
\label{subsection:github}

In addition to the official code of BERT by Google and GPT-2 by OpenAI, our implementation leverages the following repositories:

\smallskip
\begin{enumerate}[(a)]
\item imcaspar/gpt2-ml~\cite{GPT2-ML}  
\item huggingface/transformers~\cite{Wolf2019HuggingFacesTS}
\item ConnorJL/GPT2~\cite{connorjl_gpt2}
\item huggingface/tokenizers~\cite{tokenizers} 
\item hanxiao/bert-as-service~\cite{xiao2018bertservice}
\item spotify/annoy~\cite{annoy0001}
\end{enumerate} 
\smallskip

According to~\cite{jiehsheng05}, OpenAI trained
their models with TPU, but the code for training was not released. The authors in~\cite{jiehsheng03} resorted to~\cite{connorjl_gpt2} since it can leverage TPU and the trained model is compatible with OpenAI’s code for inferencing on GPU. According to~\cite{connorjl_gpt2}, a potential downside is that the performance of the 1.5B model seems inferior to the official model performance by OpenAI. Therefore, we checked alternatives and found ``transformers''~\cite{Wolf2019HuggingFacesTS} and ``gpt2-ml''~\cite{GPT2-ML}. The former is a more promising codebase for several technical reasons (omitted here for brevity). Unfortunately, we tried and realized that PyTorch's support for TPU is maturing, but the specific code for GPT-2 training is not ready. Therefore, we opted for ``gpt2-ml'' which has successfully built a 1.5B model. The ``gpt2-ml'' repository is forked from Grover~\cite{zellers2019grover}, which was developed by the Allen Institute. Grover is designed for fake news detection. According to~\cite{zellers2019grover}, Grover obtains over 92\% accuracy at telling apart human-written from machine-written news. The authors also released the 1.5B Grover GPT-2 model. The 1.5B model's availability from a reputable institute is the main reason we select the ``gpt2-ml'' repository to work on. One disadvantage of Grover's model is that it is not compatible with OpenAI's GPT-2 model. It means that we can not re-use OpenAI's code for inferencing. We have to use the inferencing code from Grover's code. We expect that ``transformers'' might be the best choice for researchers to pre-train OpenAI GPT models with TPU and retain the compatibility with OpenAI GPT-2 and GPT-3 models in the near future. Regarding the other repositories on the above list, their respective functions are: ``tokenizers''~\cite{tokenizers} for fast tokenization (replacing Google's and OpenAI's code) and building vocabulary from patent corpus, ``bert-as-service''~\cite{xiao2018bertservice} for fast conversion from text to BERT embeddings, and ``annoy''~\cite{annoy0001} for searching and ranking BERT embeddings efficiently.

\subsection{Implementation details: GPT-2}
\label{subsection:gpt2_implementation}

Before pre-training, we use ``tokenizers'' (ByteLevelBPETokenizer) to build the vocabulary specific to our patent corpus, instead of using the default vocabulary released in ``gpt2-ml''. We set the same vocabulary size (50257) to build our vocabulary. One advantage of building our own vocabulary file is that each special tag in our design can be encoded as one token instead of multiple (if using the original vocabulary by others). The model sizes we experiment with are \textit{Base} (similar to OpenAI's 117M) and \textit{Large} (similar to OpenAI's 345M).

\begin{figure}[t]
    \centering
    \includegraphics[width=\linewidth]{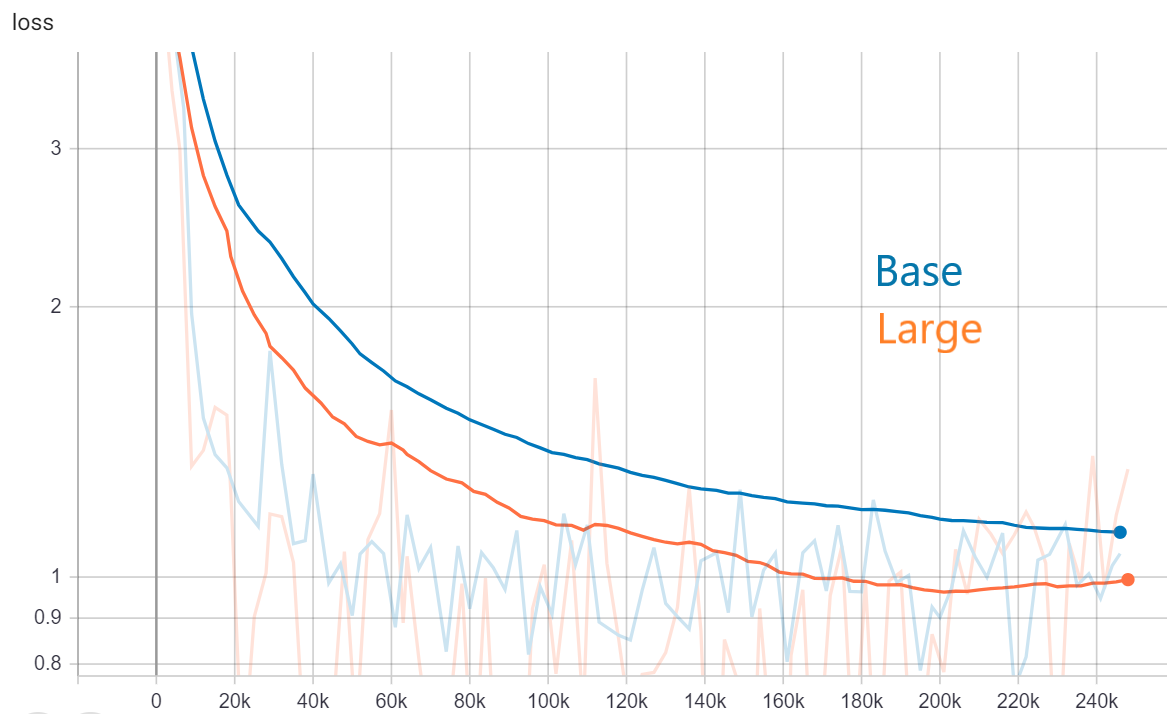}
    \caption{Training loss of GPT-2 models (Base \& Large)}
    \label{fig:gpt2_loss}
\end{figure}

The total number of tokens in the TFRecords for GPT-2 is about 32.3B. For training the \textit{Base} model, we found that batch\_size\_per\_core = 16 and \verb!max_seq_length = 1024! are workable on Colab. Larger batch size will trigger an OOM (out-of-memory) error. The number of TPU cores on Colab is 8.
Our goal is to train at least one epoch. Therefore, we set our training steps as 248,000 (32,398,927,872 / 1024 / 16 / 8 = 247,184). For training the \textit{Large} model, we set \verb!batch_size_per_core=4! to avoid the OOM error and set the same training steps. Fig.~\ref{fig:gpt2_loss} shows the curves of training loss. The final loss values are 1.122 (\textit{Base}) and 0.9934 (\textit{Large}), respectively. It is noted that the largest model (1.5B) will trigger the OOM error even after setting the batch size as 1. We leave the 1.5B model to the future when having more resources. 

\subsection{Implementation details: BERT}
\label{subsection:bert_implementation}

Before pre-training, we use ``tokenizers'' (BertWordPieceTokenizer) to build the vocabulary (uncased) specific to our patent corpus instead of using the default vocabulary released in BERT official code. We set the same vocabulary size (30522) to build our vocabulary. As for the BERT model size, we experiment with BERT-Base and BERT-Large.

According to~\cite{bert_paper}, the pre-training for the BERT-Large model took 1,000,000 steps, which is approximately 40 epochs over the 3.3 billion word corpus by using 16 Cloud TPU devices (256 batch size * 512 tokens = 128,000 tokens/batch). 
Our pre-training data contains 6.8 billion words (6,824,071,153). Since the Colab provides one Cloud TPU device only, the total number of tokens per batch is more limited (64 batch size * 128 tokens = 8,192 tokens/batch). We set our training steps as 2,000,000 to pre-train approximately 2.4 epochs over the 6.8 billion word corpus. 
Except for these, we use the same hyperparatmers provided in~\cite{bert_paper} for the BERT-Base and BERT-Large models. 
For evaluatoin, we set the \verb!eval_batch_size=32! and \verb!max_eval_steps=100,000!. The evaluation results are: 

\smallskip
\begin{itemize}
\item loss = 1.0650321
\item masked\_lm\_accuracy = 0.78279483
\item masked\_lm\_loss = 0.96379614
\item next\_sentence\_accuracy = 0.9975
\item next\_sentence\_loss = 0.0040773232
\end{itemize}
\smallskip

\begin{figure}[t]
    \centering
    \includegraphics[width=\linewidth]{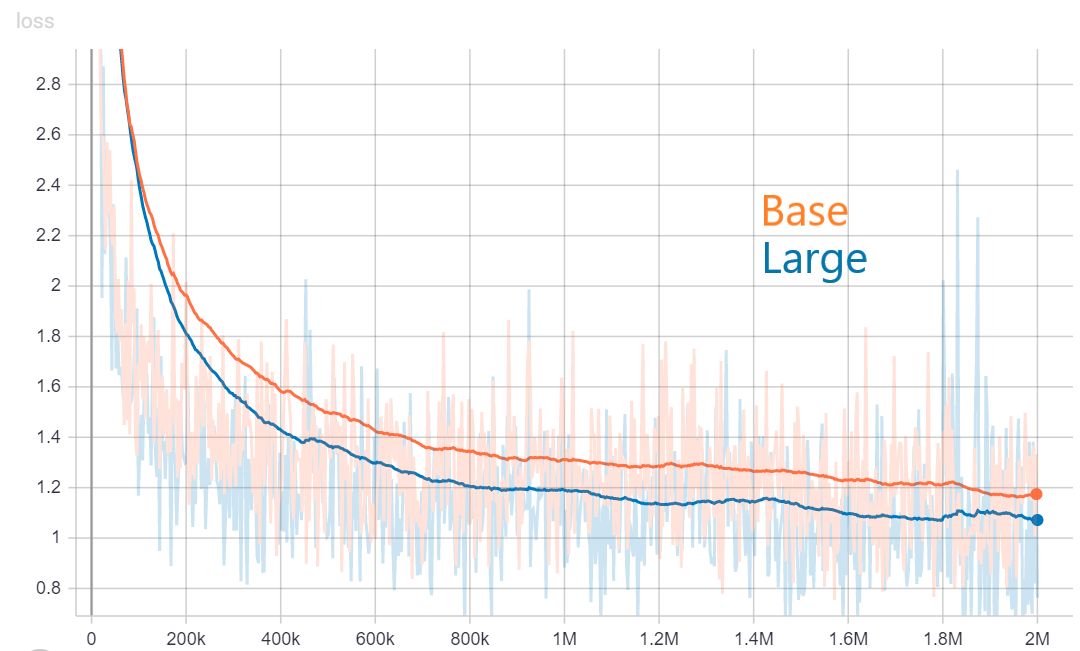}
    \caption{Training loss of BERT models (Base \& Large)}
    \label{fig:bert_loss}
\end{figure}

For comparing model performance, we trained the BERT-Base model with similar settings. Fig.~\ref{fig:bert_loss} shows the curves of training loss for the BERT-Large and BERT-Base models. As expected, the BERT-Large model has a lower curve. 

\subsection{Qualitative examples}

In this section, we provide positive and negative examples in our reranking experiments. Our proof-of-concept results (POC 1\~{}7) are available on the web.\footnote{https://usptg.herokuapp.com/mlld} In POC 1, the results contain 100 generated patent spans (no cherrypicking) in patent abstract (similar experiments can be conducted on patent claims in the future). The input for GPT-2 is the first sentence in the abstract of the US Patent 10,694,449 (granted on 2020-06-23). We selected three generated patent spans for prior art search and reranking, as below.

\smallskip
\begin{itemize}   
\item (POC 1) 
\item input: An apparatus and methods are provided for automatically detecting and connecting to a Wi-Fi network. 
\item output: [1-4] In accordance with a signal strength measurement from a Wi-Fi transceiver during an idle period when a Wi-Fi network is detected, the Wi-Fi transceiver sends on to a server an indication of a connection mode of the user equipment.
\item output: [1-33] In an embodiment, a method is provided for automatically connecting a mobile telephone to a Wi-Fi network.
\item output: [1-42] The apparatus can include a device for detecting whether a wireless device is in proximity to a wireless device associated with the Wi-Fi network.
\end{itemize}
\smallskip

Taking [1-33] (generated by GPT-2) as the input, our prior art search retrieves the top 100 records by BM25 and rerank them by embeddings. The [3/100] record in POC 4 (as below) is subjectively a positive example for us. Compared with other records in POC 4, the [3/100] record ``automatic connectivity....a mobile device to roam'' is more relevant to the ``automatically connecting a mobile telephone'' in [1-33] of POC 1. The [3/100] record is ranked as 26 based on BM25 and reranked as 3 based on embedding. Therefore, the reranking is effective in boosting its ranking. The [3/100] record is the 5th span in the abstract of patent 8590023, which was in the dataset for pre-training GPT-2 in the first place. 

\smallskip
\begin{itemize}    
\item (POC 4)    
\item patent: 8590023 [ A-4 ] (5th span in abstract)
\item text: This automatic connectivity may allow a mobile device to roam across Wi-Fi hotspots of Wi-Fi networks and offload traffic to Wi-Fi networks.
\item ranked by BM25: 26 
\item re-ranked by embedding: 3 
\end{itemize}
\smallskip

The rankings by BM25 and embedding similarity may be different or similar or the same. For example, in POC 4, the [1/100] record (as below) shows that both ranks are top 1. The [1/100] record in POC 4 is also semantically similar to the [1-33] record in POC 1. The [1/100] record in POC 4 is the first span in the abstract of patent 10356696, which was in the dataset for pre-training GPT-2 in the first place.  

\smallskip
\begin{itemize}
\item (POC 4)
\item patent: 10356696 [ A-0 ] (1st span in abstract)
\item text: An apparatus and methods are provided for automatically detecting and connecting to a Wi-Fi network.
\item ranked by BM25: 1 
\item re-ranked by embedding: 1 
\end{itemize}
\smallskip

We also found negative examples. In POC 5, the following is the top record according to both BM25 and embeddings. However, the similarity between the top record in POC 5 and the input of POC 5 (the GPT-2 output [1-4] in POC 1) seems remote. Such a result suggests that sentence similarity is still a difficult problem. One possible reason is that the coverage of the recall by BM25 is not broad enough. Therefore, it filters out suitable candidates for calculating embedding similarity too soon. 

\smallskip
\begin{itemize}
\item (POC 5)  
\item patent: 9373249 [ A-1 ] (2nd span in abstract)
\item text: The Wi-Fi transceiver receives a Wi-Fi control signal from a control signal generator.
\item ranked by BM25: 1 
\item re-ranked by embedding: 1 
\end{itemize}
\smallskip

In addition to BM25 and embedding, we found that adding a keyword can be a beneficial enhancement if a user has a clear idea about what to look for. For example, the top reranking result in the following POC 2 will be less relevant if ``proximity'' in the [1-42] record of POC 1 is the point of interest. The [1-42] record of POC 1 is the input for prior art search in POC 2. Such a result is reasonable because there is no clue for the model to weigh the point of interest more.

\smallskip
\begin{itemize}
\item (POC 2, reranked as top 1)
\item patent: 7302229 [ A-2 ] (2nd span in abstract)
\item text: In one embodiment, availability of wireless connectivity may be determined to a first user of a wireless service at a first wireless communication device to communicate with an access point associated with a Wi-Fi wireless network that offers the wireless service.
\end{itemize}
\smallskip

To boost the search relevancy, we add a keyword setting to BM25. After adding ``proximity'' as a required term in the BM25 search, in the following POC 6, the relevancy of its top record increases significantly. The total number of positive results increases too. Using a keyword as the first filter in reranking is a research topic we plan to study in the future because adding such a hard constraint could be a double-edged sword.

\smallskip
\begin{itemize}
\item (POC 6, reranked as top 1)
\item patent: 9986380 [ A-0 ] (1st span in abstract)
\item text: A first wireless device determines whether the first wireless device is in a specified proximity to a second wireless device based on a signal wirelessly transmitted by the second wireless device.
\end{itemize}
\smallskip

It is noted that POC 3 (omitted here for brevity) shows an example of a complete patent abstract containing several text spans generated by GPT-2. Each text span can go through the same prior art search with reranking, as demonstrated above. We leave such an enhancement to the future. It is also noted that, in our early experiments, using embeddings alone (without BM25) produces many false-positive results, as shown in POC 7. For example, the similarity between ``Coherent LADAR using intra-pixel quadrature detection'' and ``In-pixel correlated double sampling with fold-over detection'' is a negative example. There are many negative results with unreasonable similarities. Therefore, embeddings alone are not effective for semantic search. Comparing our initial experiments (embedding only) and later experiments (reranking by BM25 and embeddings), we conclude that the reranking is more effective even though it still produces very mixed results.  

\subsection{Failure case: few-shot learning}

Although this work focuses on GPT-2, we are also interested in the capabilities of the latest GPT-3. GPT-3 is an autoregressive language model with 175 billion parameters. According to the authors, it is 10x more than any previous language model. By scaling up, the model can perform few-shot learning purely via text interaction without any gradient updates or fine-tuning. We estimate that the largest GPT-3 model is about 507 times bigger than the GPT-2 model we utilized. We hypothesize that the patent text structure is more uniform and less diverse than the training data for GPT-3.  Hence, we wonder whether few-shot learning might be possible on our GPT-2 model too. We prepare our input text in the following format:

\smallskip
$<$\textbar start\_of\_figure\textbar$>$ (text1) 
$<$\textbar end\_of\_figure\textbar$>$
$<$\textbar figure2title\textbar$>$
$<$\textbar start\_of\_title\textbar$>$ (text2)  
$<$\textbar end\_of\_title\textbar$>$
\smallskip

The $<$\textbar figure2title\textbar$>$ mapping is defined in the vocabulary file, and there is no training data contains such a mapping. Our purpose is to test whether the model can learn such a new mapping by few-shot learning. In our experiments, we concatenate several records of different figure text and title. Then, we remove the title in the last record. If the few-shot learning works, the model should generate the removed patent title in the last record. Unfortunately, we found it not workable. The limitation in the model size is probably the primary root cause. Although such a failure case was anticipated, we found one intriguing pattern: the model keeps generating the patent title in the second record most of the time. We leave this case to the future. Determining the minimal model size to achieve few-shot learning in the patent domain is also an important topic for future research.

\section{Future research}

Our experiments show mixed results, and the topics for future researchers include: 

\smallskip
\begin{itemize}
\item How to make reranking more effective?
\item How to measure the ``novelty'' and ``non-obviousness'' (requirements in patent laws) between the generated patent text and prior patent text?
\item What are the legal \& ethical considerations before releasing a generative patent model?
\item Can the discrepancy of the rankings between BM25 and embedding be a source for data augmentation? For example, Sentence-BERT requires both positive and negative examples to train. Ranking by embeddings first and filtering by BM25 later might be a way to collect negative training examples.
\end{itemize}
\smallskip




\section{Conclusion}

Reranking with BM25 and embeddings is a practical approach for producing better search results than using embeddings alone. Our reranking is a two-step approach in which the search is performed based on BM25 first and then performed based on the cosine similarity of embeddings. If a user has a clear point of interest in mind, the search can be more productive by adding an extra step of providing a keyword to the BM25 search. In this work, the input for the prior art search is the patent text span generated by a GPT-2 model. The objective of our prior art search is to identify retrospectively the most similar patent text spans in the training data of the GPT-2 model. Although our experiments show the effectiveness of reranking in the patent domain, they also show that semantic search for longer text remains challenging. By finding the similarity between GPT-2's inputs and outputs, we expect that this work and its future enhancement can help researchers understand GPT-2 better. Particularly, in the patent domain, it is critical to evaluate the novelty in GPT-2 and GPT-3 models. 
To evaluate the novelty, a prerequisite is to identify the closest training data. 
The progress in this paper is toward fulfilling the prerequisite so that novelty of the generated patent text can be evaluated in the future.
In our system architecture, we integrate several building blocks, notably pre-training GPT-2, pre-training BERT, using Elasticsearch for BM25 ranking, and reranking by embedding similarity with Annoy. Such a proof-of-concept implementation is a practical reference for future researchers.  



\bibliographystyle{ACM-Reference-Format}
\bibliography{mybibfile}


\begin{thebibliography}{30}


\ifx \showCODEN    \undefined \def \showCODEN     #1{\unskip}     \fi
\ifx \showDOI      \undefined \def \showDOI       #1{#1}\fi
\ifx \showISBNx    \undefined \def \showISBNx     #1{\unskip}     \fi
\ifx \showISBNxiii \undefined \def \showISBNxiii  #1{\unskip}     \fi
\ifx \showISSN     \undefined \def \showISSN      #1{\unskip}     \fi
\ifx \showLCCN     \undefined \def \showLCCN      #1{\unskip}     \fi
\ifx \shownote     \undefined \def \shownote      #1{#1}          \fi
\ifx \showarticletitle \undefined \def \showarticletitle #1{#1}   \fi
\ifx \showURL      \undefined \def \showURL       {\relax}        \fi
\providecommand\bibfield[2]{#2}
\providecommand\bibinfo[2]{#2}
\providecommand\natexlab[1]{#1}
\providecommand\showeprint[2][]{arXiv:#2}

\bibitem[\protect\citeauthoryear{Brown, Mann, Ryder, Subbiah, Kaplan, Dhariwal,
  Neelakantan, Shyam, Sastry, Askell, Agarwal, Herbert-Voss, Krueger, Henighan,
  Child, Ramesh, Ziegler, Wu, Winter, Hesse, Chen, Sigler, Litwin, Gray, Chess,
  Clark, Berner, McCandlish, Radford, Sutskever, and Amodei}{Brown
  et~al\mbox{.}}{2020}]%
        {gpt-3}
\bibfield{author}{\bibinfo{person}{Tom~B. Brown}, \bibinfo{person}{Benjamin
  Mann}, \bibinfo{person}{Nick Ryder}, \bibinfo{person}{Melanie Subbiah},
  \bibinfo{person}{Jared Kaplan}, \bibinfo{person}{Prafulla Dhariwal},
  \bibinfo{person}{Arvind Neelakantan}, \bibinfo{person}{Pranav Shyam},
  \bibinfo{person}{Girish Sastry}, \bibinfo{person}{Amanda Askell},
  \bibinfo{person}{Sandhini Agarwal}, \bibinfo{person}{Ariel Herbert-Voss},
  \bibinfo{person}{Gretchen Krueger}, \bibinfo{person}{Tom Henighan},
  \bibinfo{person}{Rewon Child}, \bibinfo{person}{Aditya Ramesh},
  \bibinfo{person}{Daniel~M. Ziegler}, \bibinfo{person}{Jeffrey Wu},
  \bibinfo{person}{Clemens Winter}, \bibinfo{person}{Christopher Hesse},
  \bibinfo{person}{Mark Chen}, \bibinfo{person}{Eric Sigler},
  \bibinfo{person}{Mateusz Litwin}, \bibinfo{person}{Scott Gray},
  \bibinfo{person}{Benjamin Chess}, \bibinfo{person}{Jack Clark},
  \bibinfo{person}{Christopher Berner}, \bibinfo{person}{Sam McCandlish},
  \bibinfo{person}{Alec Radford}, \bibinfo{person}{Ilya Sutskever}, {and}
  \bibinfo{person}{Dario Amodei}.} \bibinfo{year}{2020}\natexlab{}.
\newblock \showarticletitle{Language Models are Few-Shot Learners}.
\newblock \bibinfo{journal}{\emph{ArXiv}} (\bibinfo{year}{2020}).
\newblock
\urldef\tempurl%
\url{https://arxiv.org/abs/2005.14165}
\showURL{%
\tempurl}


\bibitem[\protect\citeauthoryear{Devlin, Chang, Lee, and Toutanova}{Devlin
  et~al\mbox{.}}{2019a}]%
        {devlin-etal-2019-bert}
\bibfield{author}{\bibinfo{person}{Jacob Devlin}, \bibinfo{person}{Ming-Wei
  Chang}, \bibinfo{person}{Kenton Lee}, {and} \bibinfo{person}{Kristina
  Toutanova}.} \bibinfo{year}{2019}\natexlab{a}.
\newblock \showarticletitle{{BERT}: Pre-training of Deep Bidirectional
  Transformers for Language Understanding}. In
  \bibinfo{booktitle}{\emph{Proceedings of the 2019 Conference of the North
  {A}merican Chapter of the Association for Computational Linguistics: Human
  Language Technologies, Volume 1 (Long and Short Papers)}}.
  \bibinfo{publisher}{Association for Computational Linguistics},
  \bibinfo{address}{Minneapolis, Minnesota}, \bibinfo{pages}{4171--4186}.
\newblock
\urldef\tempurl%
\url{https://doi.org/10.18653/v1/N19-1423}
\showDOI{\tempurl}


\bibitem[\protect\citeauthoryear{Devlin, Chang, Lee, and Toutanova}{Devlin
  et~al\mbox{.}}{2019b}]%
        {bert_paper}
\bibfield{author}{\bibinfo{person}{Jacob Devlin}, \bibinfo{person}{Ming-Wei
  Chang}, \bibinfo{person}{Kenton Lee}, {and} \bibinfo{person}{Kristina
  Toutanova}.} \bibinfo{year}{2019}\natexlab{b}.
\newblock \showarticletitle{{BERT}: Pre-training of Deep Bidirectional
  Transformers for Language Understanding}. In
  \bibinfo{booktitle}{\emph{Proceedings of the 2019 Conference of the North
  {A}merican Chapter of the Association for Computational Linguistics: Human
  Language Technologies, Volume 1 (Long and Short Papers)}}.
  \bibinfo{publisher}{Association for Computational Linguistics},
  \bibinfo{address}{Minneapolis, Minnesota}, \bibinfo{pages}{4171--4186}.
\newblock
\urldef\tempurl%
\url{https://doi.org/10.18653/v1/N19-1423}
\showDOI{\tempurl}


\bibitem[\protect\citeauthoryear{D’hondt, Verberne, Weber, Koster, and
  Boves}{D’hondt et~al\mbox{.}}{2012}]%
        {Dhondt_Verberne_Weber_Koster_Boves_2012}
\bibfield{author}{\bibinfo{person}{Eva D’hondt}, \bibinfo{person}{Suzan
  Verberne}, \bibinfo{person}{Niklas Weber}, \bibinfo{person}{Kees Koster},
  {and} \bibinfo{person}{Lou Boves}.} \bibinfo{year}{2012}\natexlab{}.
\newblock \showarticletitle{Using skipgrams and PoS-based feature selection for
  patent classification}.
\newblock \bibinfo{journal}{\emph{Computational Linguistics in the Netherlands
  Journal}}  \bibinfo{volume}{2} (\bibinfo{date}{Dec.} \bibinfo{year}{2012}),
  \bibinfo{pages}{52--70}.
\newblock
\urldef\tempurl%
\url{https://www.clips.uantwerpen.be/clinjournal/clinj/article/view/15}
\showURL{%
\tempurl}


\bibitem[\protect\citeauthoryear{Google}{Google}{[n.d.]a}]%
        {bigquery}
\bibfield{author}{\bibinfo{person}{Google}.}
  \bibinfo{year}{[n.d.]}\natexlab{a}.
\newblock \bibinfo{title}{Google Patents public datasets on BigQuery}.
\newblock
  \bibinfo{howpublished}{\url{https://console.cloud.google.com/bigquery?p=patents-public-data}}.
\newblock


\bibitem[\protect\citeauthoryear{Google}{Google}{[n.d.]b}]%
        {bert_github}
\bibfield{author}{\bibinfo{person}{Google}.}
  \bibinfo{year}{[n.d.]}\natexlab{b}.
\newblock \bibinfo{title}{google-research/bert}.
\newblock
  \bibinfo{howpublished}{\url{https://github.com/google-research/bert}}.
\newblock


\bibitem[\protect\citeauthoryear{Google}{Google}{[n.d.]c}]%
        {universal-sentence-encoder}
\bibfield{author}{\bibinfo{person}{Google}.}
  \bibinfo{year}{[n.d.]}\natexlab{c}.
\newblock \bibinfo{title}{Universal Sentence Encoder}.
\newblock
  \bibinfo{howpublished}{\url{https://tfhub.dev/google/universal-sentence-encoder/2}}.
\newblock


\bibitem[\protect\citeauthoryear{HuggingFace}{HuggingFace}{2020}]%
        {tokenizers}
\bibfield{author}{\bibinfo{person}{HuggingFace}.}
  \bibinfo{year}{2020}\natexlab{}.
\newblock \bibinfo{title}{Fast State-of-the-Art Tokenizers optimized for
  Research and Production}.
\newblock
  \bibinfo{howpublished}{\url{https://github.com/huggingface/tokenizers}}.
\newblock


\bibitem[\protect\citeauthoryear{Leahy}{Leahy}{2019}]%
        {connorjl_gpt2}
\bibfield{author}{\bibinfo{person}{Connor Leahy}.}
  \bibinfo{year}{2019}\natexlab{}.
\newblock \bibinfo{title}{An implementation of training for GPT2, supports
  TPUs}.
\newblock \bibinfo{howpublished}{\url{https://github.com/ConnorJL/GPT2}}.
\newblock


\bibitem[\protect\citeauthoryear{Lee}{Lee}{2020}]%
        {jiehsheng08}
\bibfield{author}{\bibinfo{person}{Jieh-Sheng Lee}.}
  \bibinfo{year}{2020}\natexlab{}.
\newblock \showarticletitle{Measuring and Controlling Text Generation by
  Semantic Search}. In \bibinfo{booktitle}{\emph{WWW '20: Companion Proceedings
  of the Web Conference 2020}}. \bibinfo{address}{Taipei, Taiwan},
  \bibinfo{pages}{269–273}.
\newblock
\urldef\tempurl%
\url{https://doi.org/10.1145/3366424.3382086}
\showURL{%
\tempurl}


\bibitem[\protect\citeauthoryear{Lee and Hsiang}{Lee and Hsiang}{2019}]%
        {jiehsheng02}
\bibfield{author}{\bibinfo{person}{Jieh-Sheng Lee} {and} \bibinfo{person}{Jieh
  Hsiang}.} \bibinfo{year}{2019}\natexlab{}.
\newblock \showarticletitle{Measuring Patent Claim Generation by Span
  Relevancy}. In \bibinfo{booktitle}{\emph{Proceedings of the Thirteenth
  International Workshop on Juris-informatics (JURISIN)}}.
  \bibinfo{address}{Keio University Kanagawa, Japan}.
\newblock


\bibitem[\protect\citeauthoryear{Lee and Hsiang}{Lee and Hsiang}{2020a}]%
        {jiehsheng03}
\bibfield{author}{\bibinfo{person}{Jieh-Sheng Lee} {and} \bibinfo{person}{Jieh
  Hsiang}.} \bibinfo{year}{2020}\natexlab{a}.
\newblock \showarticletitle{Patent claim generation by fine-tuning {OpenAI
  GPT-2}}.
\newblock \bibinfo{journal}{\emph{World Patent Information}}
  (\bibinfo{year}{2020}).
\newblock
\newblock
\shownote{in press.}


\bibitem[\protect\citeauthoryear{Lee and Hsiang}{Lee and Hsiang}{2020b}]%
        {jiehsheng01}
\bibfield{author}{\bibinfo{person}{Jieh-Sheng Lee} {and} \bibinfo{person}{Jieh
  Hsiang}.} \bibinfo{year}{2020}\natexlab{b}.
\newblock \showarticletitle{{PatentBERT: Patent classification with fine-tuning
  a pre-trained BERT model}}.
\newblock \bibinfo{journal}{\emph{World Patent Information}}
  \bibinfo{volume}{61}, \bibinfo{number}{101965} (\bibinfo{year}{2020}).
\newblock
\urldef\tempurl%
\url{https://doi.org/10.1016/j.wpi.2020.101965}
\showURL{%
\tempurl}


\bibitem[\protect\citeauthoryear{Lee and Hsiang}{Lee and Hsiang}{2020c}]%
        {jiehsheng05}
\bibfield{author}{\bibinfo{person}{Jieh-Sheng Lee} {and} \bibinfo{person}{Jieh
  Hsiang}.} \bibinfo{year}{2020}\natexlab{c}.
\newblock \bibinfo{title}{Patent{T}ransformer-2: Controlling Patent Text
  Generation by Structural Metadata}.  (\bibinfo{year}{2020}).
\newblock
\urldef\tempurl%
\url{https://arxiv.org/abs/2001.03708}
\showURL{%
\tempurl}


\bibitem[\protect\citeauthoryear{Liu, Ott, Goyal, Jingfei~Du, Chen, Levy,
  Lewis, Zettlemoyer, and Stoyanov}{Liu et~al\mbox{.}}{2019}]%
        {RoBERTa}
\bibfield{author}{\bibinfo{person}{Yinhan Liu}, \bibinfo{person}{Myle Ott},
  \bibinfo{person}{Naman Goyal}, \bibinfo{person}{Mandar~Joshi Jingfei~Du},
  \bibinfo{person}{Danqi Chen}, \bibinfo{person}{Omer Levy},
  \bibinfo{person}{Mike Lewis}, \bibinfo{person}{Luke Zettlemoyer}, {and}
  \bibinfo{person}{Veselin Stoyanov}.} \bibinfo{year}{2019}\natexlab{}.
\newblock \showarticletitle{RoBERTa: A Robustly Optimized BERT Pretraining
  Approach}.
\newblock \bibinfo{journal}{\emph{ArXiv}} (\bibinfo{year}{2019}).
\newblock
\urldef\tempurl%
\url{http://arxiv.org/abs/1907.11692}
\showURL{%
\tempurl}


\bibitem[\protect\citeauthoryear{Logeswaran and Lee}{Logeswaran and
  Lee}{2018}]%
        {logeswaran2018an}
\bibfield{author}{\bibinfo{person}{Lajanugen Logeswaran} {and}
  \bibinfo{person}{Honglak Lee}.} \bibinfo{year}{2018}\natexlab{}.
\newblock \showarticletitle{An efficient framework for learning sentence
  representations}. In \bibinfo{booktitle}{\emph{International Conference on
  Learning Representations}}.
\newblock
\urldef\tempurl%
\url{https://openreview.net/forum?id=rJvJXZb0W}
\showURL{%
\tempurl}


\bibitem[\protect\citeauthoryear{Mikolov, Sutskever, Chen, Corrado, and
  Dean}{Mikolov et~al\mbox{.}}{2013}]%
        {word2vec}
\bibfield{author}{\bibinfo{person}{Tomas Mikolov}, \bibinfo{person}{Ilya
  Sutskever}, \bibinfo{person}{Kai Chen}, \bibinfo{person}{Greg~S Corrado},
  {and} \bibinfo{person}{Jeff Dean}.} \bibinfo{year}{2013}\natexlab{}.
\newblock \showarticletitle{Distributed Representations of Words and Phrases
  and their Compositionality}.
\newblock In \bibinfo{booktitle}{\emph{Advances in Neural Information
  Processing Systems 26}}, \bibfield{editor}{\bibinfo{person}{C.~J.~C. Burges},
  \bibinfo{person}{L.~Bottou}, \bibinfo{person}{M.~Welling},
  \bibinfo{person}{Z.~Ghahramani}, {and} \bibinfo{person}{K.~Q. Weinberger}}
  (Eds.). \bibinfo{publisher}{Curran Associates, Inc.},
  \bibinfo{pages}{3111--3119}.
\newblock


\bibitem[\protect\citeauthoryear{Radrof, Wu, Child, Luan, Amodei, and
  Sutskever}{Radrof et~al\mbox{.}}{2018}]%
        {gpt2_Radrof01}
\bibfield{author}{\bibinfo{person}{Alec Radrof}, \bibinfo{person}{Jeffrey Wu},
  \bibinfo{person}{Rewon Child}, \bibinfo{person}{David Luan},
  \bibinfo{person}{Dario Amodei}, {and} \bibinfo{person}{Ilya Sutskever}.}
  \bibinfo{year}{2018}\natexlab{}.
\newblock \bibinfo{title}{Language Models are Unsupervised Multitask Learners}.
\newblock
\newblock


\bibitem[\protect\citeauthoryear{Reimers and Gurevych}{Reimers and
  Gurevych}{2019}]%
        {SentenceBERT}
\bibfield{author}{\bibinfo{person}{Nils Reimers} {and} \bibinfo{person}{Iryna
  Gurevych}.} \bibinfo{year}{2019}\natexlab{}.
\newblock \showarticletitle{Sentence-{BERT}: Sentence Embeddings using {Siamese
  BERT-Networks}}. In \bibinfo{booktitle}{\emph{Proceedings of the 2019
  Conference on Empirical Methods in Natural Language Processing}}.
  \bibinfo{publisher}{Association for Computational Linguistics}.
\newblock
\urldef\tempurl%
\url{http://arxiv.org/abs/1908.10084}
\showURL{%
\tempurl}


\bibitem[\protect\citeauthoryear{Risch, Alder, Hewel, and Krestel}{Risch
  et~al\mbox{.}}{2020}]%
        {risch2020patentmatch}
\bibfield{author}{\bibinfo{person}{Julian Risch}, \bibinfo{person}{Nicolas
  Alder}, \bibinfo{person}{Christoph Hewel}, {and} \bibinfo{person}{Ralf
  Krestel}.} \bibinfo{year}{2020}\natexlab{}.
\newblock \bibinfo{title}{PatentMatch: A Dataset for Matching Patent Claims \&
  Prior Art}.
\newblock
\newblock
\showeprint[arxiv]{2012.13919}~[cs.IR]


\bibitem[\protect\citeauthoryear{Risch and Krestel}{Risch and Krestel}{2019}]%
        {Risch2019DomainspecificWE}
\bibfield{author}{\bibinfo{person}{Julian Risch} {and} \bibinfo{person}{Ralf
  Krestel}.} \bibinfo{year}{2019}\natexlab{}.
\newblock \showarticletitle{Domain-specific word embeddings for patent
  classification}.
\newblock \bibinfo{journal}{\emph{Data Technol. Appl.}}  \bibinfo{volume}{53}
  (\bibinfo{year}{2019}), \bibinfo{pages}{108--122}.
\newblock


\bibitem[\protect\citeauthoryear{Spotify}{Spotify}{2018}]%
        {annoy0001}
\bibfield{author}{\bibinfo{person}{Spotify}.} \bibinfo{year}{2018}\natexlab{}.
\newblock \bibinfo{title}{Approximate Nearest Neighbors in C++/Python optimized
  for memory usage and loading/saving to disk}.
\newblock \bibinfo{howpublished}{\url{https://github.com/spotify/annoy}}.
\newblock


\bibitem[\protect\citeauthoryear{Thienes and Pertschuk}{Thienes and
  Pertschuk}{2019}]%
        {nboost}
\bibfield{author}{\bibinfo{person}{Cole Thienes} {and} \bibinfo{person}{Jack
  Pertschuk}.} \bibinfo{year}{2019}\natexlab{}.
\newblock \bibinfo{title}{NBoost: Neural Boosting Search Results}.
\newblock \bibinfo{howpublished}{\url{https://github.com/koursaros-ai/nboost}}.
\newblock


\bibitem[\protect\citeauthoryear{USPTO}{USPTO}{[n.d.]a}]%
        {opendataportal}
\bibfield{author}{\bibinfo{person}{USPTO}.} \bibinfo{year}{[n.d.]}\natexlab{a}.
\newblock \bibinfo{title}{USPTO Open Data Portal}.
\newblock \bibinfo{howpublished}{\url{https://developer.uspto.gov/data}}.
\newblock


\bibitem[\protect\citeauthoryear{USPTO}{USPTO}{[n.d.]b}]%
        {patentsview}
\bibfield{author}{\bibinfo{person}{USPTO}.} \bibinfo{year}{[n.d.]}\natexlab{b}.
\newblock \bibinfo{title}{USPTO PatentsView}.
\newblock \bibinfo{howpublished}{\url{https://www.patentsview.org/download}}.
\newblock


\bibitem[\protect\citeauthoryear{Vaswani, Shazeer, Parmar, Uszkoreit, Jones,
  Gomez, Kaiser, and Polosukhin}{Vaswani et~al\mbox{.}}{2017}]%
        {Vaswani01}
\bibfield{author}{\bibinfo{person}{Ashish Vaswani}, \bibinfo{person}{Noam
  Shazeer}, \bibinfo{person}{Niki Parmar}, \bibinfo{person}{Jakob Uszkoreit},
  \bibinfo{person}{Llion Jones}, \bibinfo{person}{Aidan~N Gomez},
  \bibinfo{person}{\L~ukasz Kaiser}, {and} \bibinfo{person}{Illia Polosukhin}.}
  \bibinfo{year}{2017}\natexlab{}.
\newblock \showarticletitle{Attention is All you Need}.
\newblock In \bibinfo{booktitle}{\emph{Advances in Neural Information
  Processing Systems 30}}, \bibfield{editor}{\bibinfo{person}{I.~Guyon},
  \bibinfo{person}{U.~V. Luxburg}, \bibinfo{person}{S.~Bengio},
  \bibinfo{person}{H.~Wallach}, \bibinfo{person}{R.~Fergus},
  \bibinfo{person}{S.~Vishwanathan}, {and} \bibinfo{person}{R.~Garnett}}
  (Eds.). \bibinfo{publisher}{Curran Associates, Inc.},
  \bibinfo{pages}{5998--6008}.
\newblock
\urldef\tempurl%
\url{http://papers.nips.cc/paper/7181-attention-is-all-you-need.pdf}
\showURL{%
\tempurl}


\bibitem[\protect\citeauthoryear{Wolf, Debut, Sanh, Chaumond, Delangue, Moi,
  Cistac, Rault, Louf, Funtowicz, and Brew}{Wolf et~al\mbox{.}}{2019}]%
        {Wolf2019HuggingFacesTS}
\bibfield{author}{\bibinfo{person}{Thomas Wolf}, \bibinfo{person}{Lysandre
  Debut}, \bibinfo{person}{Victor Sanh}, \bibinfo{person}{Julien Chaumond},
  \bibinfo{person}{Clement Delangue}, \bibinfo{person}{Anthony Moi},
  \bibinfo{person}{Pierric Cistac}, \bibinfo{person}{Tim Rault},
  \bibinfo{person}{R'emi Louf}, \bibinfo{person}{Morgan Funtowicz}, {and}
  \bibinfo{person}{Jamie Brew}.} \bibinfo{year}{2019}\natexlab{}.
\newblock \showarticletitle{HuggingFace's Transformers: State-of-the-art
  Natural Language Processing}.
\newblock \bibinfo{journal}{\emph{ArXiv}} (\bibinfo{year}{2019}).
\newblock
\urldef\tempurl%
\url{https://arxiv.org/abs/1910.03771}
\showURL{%
\tempurl}


\bibitem[\protect\citeauthoryear{Xiao}{Xiao}{2018}]%
        {xiao2018bertservice}
\bibfield{author}{\bibinfo{person}{Han Xiao}.} \bibinfo{year}{2018}\natexlab{}.
\newblock \bibinfo{title}{bert-as-service}.
\newblock
  \bibinfo{howpublished}{\url{https://github.com/hanxiao/bert-as-service}}.
\newblock


\bibitem[\protect\citeauthoryear{Zellers, Holtzman, Rashkin, Bisk, Farhadi,
  Roesner, and Choi}{Zellers et~al\mbox{.}}{2019}]%
        {zellers2019grover}
\bibfield{author}{\bibinfo{person}{Rowan Zellers}, \bibinfo{person}{Ari
  Holtzman}, \bibinfo{person}{Hannah Rashkin}, \bibinfo{person}{Yonatan Bisk},
  \bibinfo{person}{Ali Farhadi}, \bibinfo{person}{Franziska Roesner}, {and}
  \bibinfo{person}{Yejin Choi}.} \bibinfo{year}{2019}\natexlab{}.
\newblock \showarticletitle{Defending Against Neural Fake News}. In
  \bibinfo{booktitle}{\emph{Advances in Neural Information Processing Systems
  32}}.
\newblock


\bibitem[\protect\citeauthoryear{Zhang}{Zhang}{2019}]%
        {GPT2-ML}
\bibfield{author}{\bibinfo{person}{Zhibo Zhang}.}
  \bibinfo{year}{2019}\natexlab{}.
\newblock \bibinfo{title}{GPT2-ML: GPT-2 for Multiple Languages}.
\newblock \bibinfo{howpublished}{\url{https://github.com/imcaspar/gpt2-ml}}.
\newblock


\end{thebibliography}




\end{document}